\documentclass[10pt]{paper}
\begin{document}

\title{Textual Entailment Recognizing  by Theorem Proving Approach}

\author{
  Doina Tatar  and Militon Frentiu\\
Department of Computer Science\\
University "Babes-Bolyai", Cluj-Napoca, Romania}

\date{}

\maketitle
\begin{abstract}

   In this paper we present two original methods for recognizing
   textual inference.First one  is a modified  resolution method
   such that some linguistic considerations are introduced in the
   unification of two atoms. The approach is possible due to the
   recent methods of transforming texts in logic formulas. Second
   one is based on semantic relations in text, as presented in WordNet.
   Some similarities between these two methods are remarked.

 {\bf Key words}: unification, resolution, textual inference, WordNet.

\end{abstract}

\section{ Introduction}

    The recognition of textual inference is one of the most complex task in Natural Language
   Understanding. Thus, a very important problem in some computational
   linguistic applications (as Question Answering, summarization, segmentation of discourse,
   coherence and cohesion  of a discourse and others) is to establish if a sentence
   {\it follows} from  a text. That means
   in many application it is important to establish if some sentences which are not existing in text
   are "logical" implied (can be inferred) by this text. The importance of text inference
    in computational linguistic is proved by the existence of
   RTE conferences (Recognizing Textual Entailment,
  http:// www.pascal-network.org/ Challenges/ RTE/), with the
  forth edition this year,
   where the main  task is   to establish the textual entailment  relation.
  For RTE-1 contest  data set includes 1367 English $T , H $  pairs
  (567 for training stage in learning methods and 800 for test).  Here the task is to determine if the meaning
  of one text (the entailed hypothesis, $H$) can be inferred from the meaning
  of the other text (the entailing text, $T$).

   On the other hand is well known that a linguistic text can be represented by a set of logical formulas,
   called logic forms. Various method were  given for associating a logical formula
   with a text:  \cite{harabagiu1},
   \cite{vasilerus}, \cite{Thayse}, \cite{bosmarkert}.  From logical point
    of view,
   if each sentence is represented as a formula, proving a textual inference consists in showing that
 a logical formula is deducible from a set of others formulas. The problem
 is for First Order Logic
  a classical (semidecidable) problem. In last years, when text mining
  is very important in many
  AI applications, text inference from both point of view,
  theorem proving and from linguistics perspective, is a very active field of research.
      In \cite{tatumoldovan} is presented a system participating
   in the RTE-1 competition, using some   world knowledge axioms and  a  theorem proving
   tool.

  Let us denote entailment relation  between a text $T$ and a  sentence or a group
  of sentences $H$ by  $ T \Rightarrow H$.

In this paper we propose two method to  solve the problem of establishing if $ T \Rightarrow H$
 : first is obtained from the classical resolution refutation method,
 completing the unification of two atoms with some  linguistic considerations
 (Modified  Resolution Method or MRM).
 Our method differs of \cite{raina} by the fact that it does  not
 need
 learning stage and it does not need  a graph representation and evaluation.
 The weight (cost) of a deduction is obtained only from the weights (costs)
 of each modified resolution steps. At his turn, the cost of a step of resolution
 is obtained by similarity considerations using some linguistic tools as
 WordNet \cite{wordnet} and Word::Similarity \cite{Pedersen}.
 No background knowledge \cite{bosmarkert} is needed.

 The second method is based on  lexical chains (paths) for entailment
  spanning the text $T$ and the text $H$
 ( Lexical-chains Based Method or LBM). A system of rules for construction of
 lexical rules corresponding to entailment is established. We claim
  that MRM and LBM   produce similar results.

In section 2 we will define our  modified unification   of two
atoms method, our lexical resolution rule and  modified resolution
method (MRM).

 In section 3 we will describe LBM method and  we will
propose another definition for text inference based on the cohesion of texts.

\section{Text inference as theorem proving.}

   Consider a knowledge base formed by a set of natural language sentences, $K$. Let define
    a set of inferences rules which is sound, in the sense that it derive
    true new sentences when the initial sentences in $K$ are true.
  It is a long debate about  formalisms to represent knowledge such that above desiderata
  be fulfilled \cite{Thayse}. We will use here the method proposed by \cite{vasilerus}
   of obtaining logical forms (in fact, logical formulas) from sentences
   expressed in natural language. In this method  each {\it open class} word in a sentence
   (that means noun, verb, adjective, adverb) is  transformed in a logic predicate (atom).
    We consider, additionally, that the constants are denoted by the names of words they
    represent.
    For these atoms  we propose a new algorithm for unification  which modifies  the
    classical Robinson unification algorithm  by adding some lexical relaxations.
   The   semantic information is used in the way we define unification
  between two atoms, as described in the following section.







\subsection{ Lexical unification method for two atoms}

       Lexical unification method of two atoms supposes that we have a lexical knowledge base
where the similarity between two words is quantified. A such of
lexical knowledge base is WordNet \cite{wordnet}, a lexical
resource which, from his construction in 1998 at
 Princeton University, is largely used in many linguistic application. Moreover,
 some connected resources are constructed (also free) which make use of WordNet
 easier. For example, Word::similarity  is an on-line interface
  which calculates the similarity between two words using some different similarity measure,
  all these starting from WordNet facilities \cite{Pedersen1},
  http://www.d.umn.edu/~tpederse/similarity.html.
  It offers the possibility to calculate
similarity between two words, two words annotated with POS, or even two words
annotated with POS and sense (in WordNet notation). Measures used to calculate
similarity could be nine, the most well known are Path lenght,
Leacok and Chodorow, Wu and Palmer and  Resnik \cite{wordnet}.
Of course, a maximal similarity is between words belonging to the same
synset (concept).

  In the following algorithm we consider that each word of a natural language sentence
  is transformed in atom as in \cite{vasilerus}. See our section 2.3. It can be seen
  there that
    the terms (the arguments) of an atom are always {\it variables or constants}. The classical
  unification of atoms is replaced by {\it lexical unification}, which depends on the
  similarity in the dictionary WordNet and which relies on this
  remark. Thus, two terms are unifiable if: they are equal,  they
  are words in the same Wordnet synset or their similarity as
  words is bigger than a threshold.

In the following algorithm we consider
  that $sim(p,p')$ between two words $p,p'$  is that obtained by the Word::similarity
  interface.


\vspace{1cm}
\hspace{1mm}  {\bf INPUT}: Two atoms $a=p(t_1,...,t_n)$ and $a'=p'(t'_1,...,t'_m)$,
 $n \leq m$, threshold $\tau$, threshold for a step $\tau'$ .
  The names $p$ and $p'$ are also words in a lexical knowledge base.

\hspace{1mm}  {\bf OUTPUT}: Decision: The atoms are lexical
unifiable (with the score $W$  being  bigger than the threshold
$\tau$)
  \hspace{1cm}  and the
unificator is $\sigma$, OR they are not unifiable for the
threshold $\tau$.

\vspace{0.5cm}

\hspace{1.5cm}   Step 1. $\sigma$ = empty substitution, $W$=0.

\hspace{1.5cm}   Step 2. If $p = p'$ (similarity is maximal, =1)
or $sim(p,p') \geq \tau'$

\hspace{2.7cm}                             then  $W:=W + sim(p,p')$ ; go to Step 3

 \hspace{2.7cm}                            else STOP: " $a$ and $a'$ are not lexical
                                         unifiable"

\hspace{1.5cm}  Step 3. If (for each $t_i, i=1,...,n$ exists $t'_j$ in $\{t'_1,...,t'_m\}$
such that
 $t_i$ and $t'_j$ are lexical unifiable with the score $W'$, $W'> \tau'$ and the composition of all unificators is
  $\sigma'$  {\bf OR} for each $t'_j, j=1,...,m$ exists $t_i$ in $\{t_1,...,t_n\}$ such that
 $t_i$ and $t'_j$ are lexical unifiable with the score $W'$ , $W'>
 \tau'$ and
 the composition of all unificators is
 $\sigma'$), and  if the new score $W$ (where each $W'$ is added )is greater  than the threshold $\tau$

\hspace{2.7cm}                then

\hspace{2.9cm}                           STOP: " $a$ and $a'$ are lexical  unifiable and

     $\sigma:= \sigma \,\, composed \,\,with \,\,  \sigma'$"

\hspace{2.7cm}                 else

\hspace{2.9cm}                         STOP: " a and a' are not lexical  unifiable"

\vspace{0.1cm}

  Let us observe that the two terms $t_i$ and $t'_j$ are lexical unifiable in the following two cases.

 1. First cases are regular cases in FOPC:

 \begin{itemize}
 \item terms are equal constants;
 \item  one is a variable, the other is a constant;
 \item  both are variables.
 \end{itemize}
  In this case the score of lexical unification is 1.

2. In the second case,  if $t_i$ and $t'_j$  are two different
constants, as they are words in KB, then they are unifiable if
$sim(t_i,t'_j) \geq \tau' $.

3. Additionally, the similarity $sim(p,p')$ is  big when $p,p'$
are from the same synset in Wordnet.

4. As a reviewer pointed up, for lexical unification {\it
kill(Oswald,Kennedy)} is unifiable with {\it
kill(Kennedy,Oswald)}. This is true, however, is hard to obtain
for the text $T$ the transcription {\it kill(Oswald,Kennedy)} and
for the hypothesis $H$ {\it kill(Kennedy,Oswald)}, when the same
tool for the translation of a text in a logical formula is used.

  The similarity between two words is  used to calculate
a score for unifiability of two atoms. The test in this case is
that the score is larger than a threshold $\tau $.  The "assumption cost model"
 presented in  \cite{manning 1} uses a similarity  measure  for  some dependency
 graphs matching.
The difference with our method is that they calculate all
unificators and choose the best one (which minimizes a given
cost). For the modified resolution method, we need to obtain the
empty clause once. The "cost" of resolution is  restricted to be
low (the score is high), while the condition of step threshold is
applied.

\subsection{Lexical resolution rule}

   The lexical resolution rule  $LR$,
 consists in considering of lexical unification of two atoms as
  replacing    regular unification:\\

{\bf  Definition}

   Two (disjunctive) clauses $c_i$ and $c_j$ provide by {\it lexical resolution}
 the (disjunctive) clause $c_k$
with the score $W$, written as

       $$ c_i , c_j \models_{ lexical \,resolution} c_k \,\,
       \,\, or\,\,\,, shortly, \,\,  c_i , c_j \models_{lr} c_k $$
  if $c_i=l \vee c_i', c_j= \neg l' \vee  c_j'$ , $l$ and $l'$ are lexical unifiable
  with the score $W$ and the unificator  $\sigma$. The resulting clause is
$c_k = \sigma(c_i')  \vee  \sigma(c_j')$.

  Remark: by disjunctive clause we mean a disjunction of literals (negated or not negated atoms).

  We will call {\it modified resolution} method
   the repeatedly application of lexical resolution rule
  . So, the {\it modified resolution} is the
  transitive closure  of the lexical resolution.

  The following definition is a translation of Robinson's definition for classical
   resolution method:\\




{\bf Definition}

 A set of disjunctive clauses $C$ obtained from formulas associated to
sentences of a text is {\it lexical  contradictory} for the
threshold $\tau $ if the empty clause [] is
   obtained from the set of formulas  $C$ by the modified resolution, and the sum of all
 scores of  lexical resolution steps (rules) is  bigger than  $\tau $:

                    $$C \models_{lr}^{*} []  $$

As in the case of  classical
  resolution, the modified resolution is a problem of search. If we
  impose in this search problem to choose each time the clauses
  with the biggest  score of lexical resolution, we obtain the empty
  clause (in the case of a set of {\it lexical  contradictory}
  clauses) with the biggest score of derivation.

Let us resume the steps of
  demonstrating by modified resolution method  that a text $T$ entails the sentence $H$
  with the weight $\tau$, property denoted by $ T \Rightarrow_{MRM,\tau} H$  :

    \begin{itemize}

    \item Translate $T$ in a set of logical formulas $T'$ and $H$ in $H'$
    (as in the following subsection).

    \item Consider the set of formulas $T' \cup neg (H')$, where by $neg (H')$
we mean the logical negation of formula $H'$

    \item Find the set $C$  of disjunctive clauses of the set of formulas  $T'$ and $neg (H')$

     \item  Verify if the set $C$ is lexical contradictory  for the threshold  $\tau $.
      In this case

                  $$ T \Rightarrow_{MR,\tau} H$$

\end{itemize}

\subsection{Logical form derivation from sentences}

 We will use the method  established by \cite{vasilerus} which is applied to texts
 which are part of speech tagged and  syntactic analyzed.

 The method is the following:

 \begin{itemize}
 \item A predicate is generated for every noun, verb, adjective and adverb (possibly
 even for prepositions and conjunctions). The name of a predicate
 is obtained from the morpheme of word;

 \item If the word is a noun, then the corresponding predicate will have as argument a variable,
 as individual object. Example: {\it person(x2)}.

 \item If the word is a verb, then the corresponding predicate will have as  first argument
 an argument  for the event (or action denoted by the verb). Moreover, if
 the verb is intransitive it will have as arguments two variables:
 one for the event  and one for the subject  argument. If the verb is transitive it will have as arguments three variables: one
 for the event, one for the subject and one for the direct  complement.
  If the verb is ditransitive it will have  as arguments four
  variables: two  for the event  and  the subject and two  for the direct complement
  and the indirect complement.

\item The arguments of verb predicates are always in the order: event, subject, direct object, indirect object.
(the condition is not necessary for  modified unification)

\item If the word is an adjective (adverb) it will introduce a predicate with the same
argument as the predicate introduced for modified noun (verb).

Example: {\it man-made object} is translated as: {\it object(x1) AND  man-made(x1)}

 \item If the  word is a preposition or a conjunction it will introduce a predicate
  with the same argument as the modified word.

  \end{itemize}

   Some transformation rules that create predicates
and assign them
arguments are presented in \cite{vasilerus}. These are obtained from
the set of rules of the syntactic analyzer. For example,
 the rule for the introduction of noun predicate is
$ ART \,\,NOUN \longrightarrow noun(x_1) $.  The rule for introduction of adverb predicate is:
$ VERB \,\,ADVERB \longrightarrow verb(e_1,x_1,x_2) \,\,AND \,\, adverb(e_1) $.

     Let us consider the following example  from \cite{tatumoldovan}:

  {\it T: John and his son, George, emigrated with Mike, John's uncle, to US in 1969}

  {\it H: George and his relative, Mike, came to America}\\

  The logical form obtained for $T$ is:

$$  John(x_1) \wedge son(x_2) \wedge George(x_2) \wedge emigrated(e_1)  \wedge Agent(x_1, e_1)$$
$$  \wedge Agent(x_2, e_1) \wedge Mike(x_3) \wedge uncle(x_1, x_3) \wedge Location(e_1, x_4)$$
$$  \wedge US(x_4) \wedge Time (e_1,x_5) \wedge 1969(x_5) $$

  The logical form obtained for $H$ is:

  $$  George(x_1) \wedge relative(x_2) \wedge Mike(x_2) \wedge came(e_1)
   \wedge Agent(x_1, e_1)$$
$$  \wedge Agent(x_2, e_1) \wedge America(x_3) \wedge  Location(e_1, x_3)
   $$

  Applying the   unification lexical method for two atoms and lexical resolution rule  for
   the obtained
  disjunctive clauses,
we obtain empty clause, as follows.

  First, the  set of clauses for $neg (H)) $ is formed by only one disjunctive clause:

  $$ \neg George(x_1) \vee \neg relative(x_2) \vee \neg Mike(x_2) \vee \neg came(e_1)
   \vee \neg Agent(x_1, e_1)$$
$$  \vee \neg Agent(x_2, e_1) \vee \neg America(x_3) \vee \neg Location(e_1, x_2)$$

   Then,  if we apply modified unification between the following pairs of atoms, the empty clause is obtained:\\

   $ relative(x_2),
    uncle(x_1, x_3)$

     $ America(x_3), US(x_4)$,

      $emigrated(e_1), came (e_1)$.

    The similarities  for the pair   $relative, uncle$, for the pair $America, US$
    and for the pair $emigrated, came$ are  calculated with Word::similarity.
     So   $ T \Rightarrow_{MRM,\tau} H$
    where  the threshold  $\tau$ must be be smaller then  the sum of these similarities.

    Let us remark that in \cite{tatumoldovan} the result is obtained using additionally
    6 axioms.

\section{Entailment on linguistic bases}

   In this section we will introduce another definition for entailment  between a text
    $T$ and a sentence $H$. This definition is based on the concept of
lexical paths and on the semantical relations presented on  WordNet.

    In the huge knowledge base which is WordNet there are many semantic relations
    which are defined between synsets of nouns,  verbs,  adverbs and of
    adjectives.  Synsets in WordNet (or concepts) are set of words which are:

    a) with the same POS and

    b) are similar as meaning (or synonyms).

    The most well known semantical relation is the
    relation {\it IS-A} between synsets of nouns (or of verbs). The
    relations {\it ENTAIL} and {\it CAUSE-TO}  defined only between synsets of
    verbs, are the most suited for purposes of entailment study.

    We will define a {\it lexical path for entailment} between two words $w_1$ and $w_2$
     , denoted by $LPE(w_1,w_2)$, a path of the form:

     $$LPE(w_1,w_2)= c_1r_1c_2r_2......r_{k-1}c_k $$

\noindent where $w_1$ is from the synset $c_1$, $w_2$ is from the synset $c_k$ and
each  relation
$r_j$ is a  semantical WordNet relation  of the form {\it IS-A} or {\it ENTAIL} or  {\it CAUSE-TO} between  synsets   $c_j$ and $c_{j+1}$.
 A {\it lexical path for entailment},     $LPE(w_1,w_2)$,  can
    be described as a regular expression of the form:

    $c_1r_1c_2r_2......r_{k-1}c_k \in ((<concept>(IS-A ))^*(<concept> (ENTAIL))^*
    \mid ((<concept>(IS-A ))^* (<concept>(CAUSE-TO)^*)^* <concept>$

       The relations {\it IS-A}, {\it ENTAIL} and   {\it CAUSE-TO}
       are transitive and no simetric.
       Thus the paths $LPE(w_1,w_2)$ and all the concepts defined
       using they have an orientation from $w_1$ to $w_2$.\\
\newpage

{\bf Definition}

      $ T \Rightarrow_{LPE,\tau} H$  if card($\{LPE(w_1,w_2)\mid
      w_1 \in T, \,\, w_2 \in H \}$) is greater than a given threshold $\tau$.

      A method to construct a path $LPE(w_1,w_2)$ is to apply the following rules:

      \begin{itemize}
    \item From $c_1 \,\, IS-A \,\,c_2$ and $c_2 \,\, IS-A \,\,c_3$ it results $c_1 \,\,
    IS-A \,\,c_3$

    \item From $c_1 \,\, IS-A \,\,c_2$ and $c_2 \,\, ENTAIL\,\,c_3$ it results $c_1 \,\,
      ENTAIL \,\,c_3$

    \item From $c_1 \,\, ENTAIL\,\, c_2$ and $c_2 \,\, IS-A \,\,c_3$ it results
    $c_1 \,\, ENTAIL \,\, c_3$

    \item From $c_1 \,\, ENTAIL\,\, c_2$ and $c_2 \,\, ENTAIL \,\,c_3$ it results
      $c_1 \,\, ENTAIL \,\,c_3$

    \item From $c_1 \,\, IS-A \,\,c_2$ and $c_2 \,\, CAUSE-TO \,\,c_3$ it results
    $c_1 \,\,CAUSE-TO \,\, c_3$

    \item From $c_1 \,\, CAUSE-TO \,\,c_2$ and $c_2 \,\, IS-A \,\,c_3$ it results
    $c_1 \,\, CAUSE-TO\,\, c_3$

    \item From $c_1 \,\, CAUSE-TO\,\, c_2$ and $c_2 \,\, CAUSE-TO \,\,c_3$ it results
      $c_1 \,\, CAUSE-TO \,\,c_3$

    \item From $c_1 \,\, CAUSE-TO  \,\,c_2$ and $c_2 \,\, ENTAIL\,\, c_3$ it results
     $c_1 \,\, ENTAIL \,\,c_3$

    \item From $c_1 \,\, ENTAIL \,\, c_2$ and $c_2 \,\,CAUSE-TO \,\, c_3$ it results
    $c_1 \,\, ENTAIL \,\,c_3$

\end{itemize}
    We claim that the following theorem holds:\\

{\bf Theorem}

     For  each given threshold $\tau$ there exists a  threshold $\tau'$ such that  the relation $ T \Rightarrow_{LPE,\tau} H$
holds iff   $ T \Rightarrow_{MRM,\tau'} H$ holds.\\

\section{Conclusions and further work}

   In this paper we presented two methods for recognizing textual inference: one is from
   the logic resolution area, using a modified unification algorithm, the second is
   a pure semantic lexical method and uses the big facilities offered by the huge
   semantical dictionary WordNet. We consider that the meaning  of these methods
  has common roots: the similarity between two atoms in unification algorithm and the
  lexical path for entailment are calculated considering semantical relations which exist
  between concepts (synsets) in WordNet. A study of the  relation between
  $\tau$, $\tau'$ is in our attention.

The combined methods in Artificial Intelligence between  approaches
   so different, as Logic and Linguistics, are very largely developed in the last time.
   The present paper belongs to this category of combined methods.

\end{document}